%% file: main-4281-goldman.tex
\documentclass[11pt,a4paper]{article}
\usepackage{times}
\usepackage{latexsym}
\usepackage{url}
\usepackage[T1]{fontenc}

\usepackage{microtype}
\usepackage{graphicx}
\usepackage{enumitem}
\usepackage{multirow}
\usepackage{amsmath}
\usepackage{caption}
\usepackage{subcaption}
\usepackage[titlenumbered,ruled]{algorithm2e}
\usepackage{mxedruli}
\usepackage{cjhebrew}
\newcommand{\kat}[1]{\begin{mxedr}#1\end{mxedr}}

\newcommand\fauxsc[1]{\fauxschelper#1 \relax\relax}
\def\fauxschelper#1 #2\relax{%
  \fauxschelphelp#1\relax\relax%
  \if\relax#2\relax\else\ \fauxschelper#2\relax\fi%
}
\def\Hscale{.85}\def\Vscale{.74}\def\Cscale{1.12}
\def\fauxschelphelp#1#2\relax{%
  \ifnum`#1>``\ifnum`#1<`\{\scalebox{\Hscale}[\Vscale]{\uppercase{#1}}\else%
    \scalebox{\Cscale}[1]{#1}\fi\else\scalebox{\Cscale}[1]{#1}\fi%
  \ifx\relax#2\relax\else\fauxschelphelp#2\relax\fi}

\usepackage[acceptedWithA]{tacl2021v1}

\title{Morphology Without Borders: Clause-Level Morphology}

\author{Omer Goldman \\
  Bar Ilan University \\
  \texttt{omer.goldman@gmail.com} \\\And
  Reut Tsarfaty \\
  Bar Ilan University \\
  \texttt{reut.tsarfaty@biu.ac.il} \\}

\begin{document}

\setlength{\abovedisplayskip}{0pt}
\setlength{\belowdisplayskip}{0pt}
\setlength{\abovedisplayshortskip}{0pt}
\setlength{\belowdisplayshortskip}{0pt}
\setlength{\jot}{0pt}%

\maketitle

\input{content}

\bibliography{anthology,custom}
\bibliographystyle{acl_natbib}

\end{document}

%% file: content.tex
\begin{abstract}
Morphological tasks 
use large multi-lingual datasets that organize {\em words} into inflection tables, which then serve as training and evaluation data for  various tasks.
However, a closer inspection of these data reveals profound cross-linguistic inconsistencies, that arise from the lack of a clear linguistic and operational definition of {\em what is a word}, and that severely impair the universality of the derived %
tasks. 
To overcome this deficiency, we propose to view morphology as a {\em clause-level} phenomenon, rather than word-level. It is anchored in a fixed yet inclusive set of features,
that encapsulates all functions realized in a saturated clause. We deliver {\sc MightyMorph}, a novel 
dataset for {\em clause-level morphology} covering 4 typologically-different languages: English, German, Turkish and Hebrew. 
We use this dataset 
to derive 3 clause-level morphological tasks: inflection, reinflection and analysis. Our experiments show that the clause-level tasks are substantially harder than the respective word-level tasks, while having comparable complexity across languages. Furthermore, redefining morphology to the clause-level provides a neat interface with contextualized language models (LMs) and allows assessing the morphological knowledge encoded in these models and their usability for morphological tasks. 
Taken together, this work opens up new horizons in the study of computational morphology, leaving ample space for studying neural morphology 
cross-linguistically.

\end{abstract}

\section{Introduction}

\begin{figure}[t]
\begin{center}
\includegraphics[width=\columnwidth]{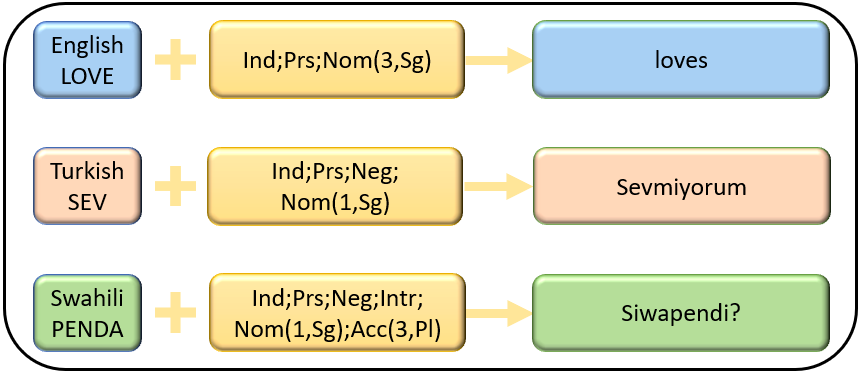}\\
\includegraphics[width=\columnwidth]{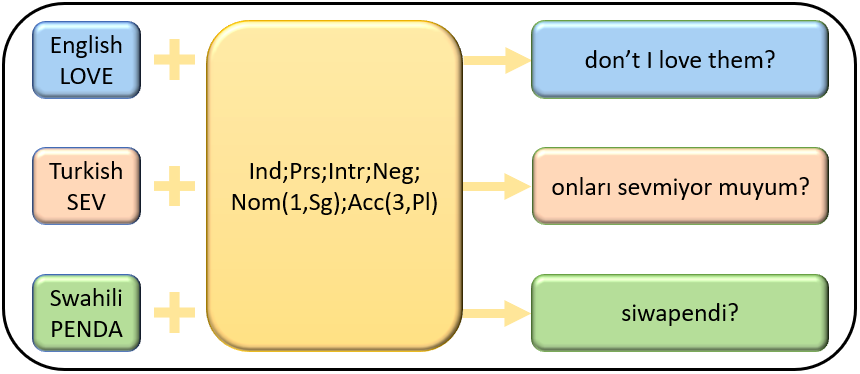}
\end{center}
\caption{In word-level morphology (top), inflection scope is defined by `wordhood', and lexemes are inflected to  different sets of features in the bundle depending on language-specific  word definitions. 
In our proposed clause-level morphology (bottom)  %
inflection scope is fixed to the same feature bundle in all languages, regardless of %
 white-spaces.}
\label{fig:overview}
\end{figure}

Morphology %
has long been viewed as a fundamental part of NLP, especially in cross-lingual settings  --- from translation \citep{minkov-etal-2007-generating, chahuneau-etal-2013-translating} to sentiment analysis \citep{abdul-mageed-etal-2011-subjectivity,amram-etal-2018-representations} --- as languages vary wildly in the extent to which they use morphological marking as a means to realize meanings. 

Recent years have seen a tremendous development in the data available for supervised morphological tasks, mostly via UniMorph \citep{batsuren-etal-2022-unimorph}, a large multi-lingual  dataset that provides morphological analyses of standalone words, organized into inflection tables %
in over 170 languages. Indeed, UniMorph was used in {all} of SIGMORPHON's shared tasks in the last decade (\citealp{cotterell-etal-2016-sigmorphon, pimentel-ryskina-etal-2021-sigmorphon} \textit{inter alia}).

Such labeled morphological data rely heavily on the notion of a {\em `word'}, as words are the elements occupying the cells of the inflection tables, and subsequently words are used as the input or output in the  morphological tasks derived from these tables.
However, a closer inspection of the data in UniMorph reveals that it is inherently inconsistent with respect to how {\em words} are defined. For instance, it is inconsistent  with regards to the inclusion or exclusion of auxiliary verbs such as "will" and "be" as part of the inflection tables, and it is inconsistent in the features words inflect for. A superficial attempt to  fix this problem leads to the can of worms that is the theoretical linguistic debate regarding the definition of \textit{the morpho-syntactic word}, where it seems that a coherent cross-lingual definition of words is nowhere to be found \cite{haspelmath-2011-indeterminacy}.

Relying on a cross-linguistically ill-defined concept in NLP is not unheard of, but it does have its price here; it undermines the perceived universality of the morphological tasks, and skews annotation efforts as well as models' accuracy in favor of those privileged languages in which morphology is not complex. To wit, even though English and Turkish exhibit comparably complex systems of tense and aspect marking, pronounced using linearly ordered morphemes, English is said to have a tiny verbal paradigm of 5 forms in UniMorph while Turkish has several hundred forms per verb.

Moreover, although inflection tables have a superficially similar structure across languages, they are in fact built upon  language-specific 
sets of features. As a result, models are tasked with {\em arbitrarily different} dimensions of meaning, guided by each language's orthographic tradition (e.g., the abundance of white-spaces used) rather than the set of functions being realized. In this work we set out to remedy such cross-linguistic inconsistencies, by delimiting the realm of morphology by the set of {\em functions} realized, rather than the set of {\em forms}.

Concretely, in this work we propose to reintroduce universality into morphological tasks by side-stepping the issue of {\em what is a word} and giving up on any attempt to determine consistent word boundaries across languages. 
Instead, we anchor morphological tasks in a cross-linguistically consistent  set of {\em inflectional  features}, that is  equivalent to a fully-saturated {\em clause}. %
Then, the lexemes in all languages are inflected to \textit{all} legal feature combinations of this set, regardless of the number of `words' or `white spaces' needed to realize its meaning. Under this revised definition, the inclusion of the
Swahili form \textit{`siwapendi'} for the lexeme {\it penda} inflected to the following features: \textsc{prs;neg;nom(1,sg);acc(3,pl)}, entails the inclusion of the English form \textit{`I don't love them'}, bearing the {\em exact same} lexeme and features.

We thus present \textsc{MightyMorph}, a novel dataset for clause-level inflectional morphology, covering 4 typologically different languages: English, German, Turkish and Hebrew. We sample data from \textsc{MightyMorph} for 3 clause-level morphological tasks: {\em inflection, reinflection} and {\em analysis}. We experiment with standard and state-of-the-art models for word-level morphological tasks \citep{silfverberg-hulden-2018-encoder, makarov-clematide-2018-neural,peters-martins-2020-one} and show that clause-level tasks are substantially harder compared to their word-level counterparts, while exhibiting comparable cross-linguistic complexity.

Operating on the clause level also neatly interfaces morphology with general-purpose pre-trained language models, such as T5 \citep{raffel-etal-2020-exploring} and BART \citep{lewis-etal-2020-bart}, to harness them for morphological tasks which were so far considered non-contextualized. Using the multilingual pre-trained model mT5  \citep{xue-etal-2021-mt5} on our data shows that complex morphology is still genuinely challenging for such LMs.
We conclude that our redefinition of morphological tasks is more theoretically sound, crosslingually more consistent, and lends itself to more sophisticated modelling, leaving ample space to  test the ability of  LMs to encode complex morphological phenomena.

The contributions of this paper are manifold. First, we uncover a major inconsistency in the current setting of supervised morphological tasks in NLP (\S\ref{sec:word_morph}). Second, we redefine  morphological inflection to the clause level (\S\ref{sec:clause_morph}) and deliver \textsc{MightyMorph}, a novel clause-level morphological  dataset  reflecting the revised definition (\S\ref{sec:mightymorph}). We then present data for 3 clause-level morphological tasks with strong baseline results  for all languages, that demonstrate the profound challenge posed by our new approach to contemporary models (\S\ref{sec:experiments}). %

\section{Morphological Essential Preliminaries}
\label{sec:word_morph}
\subsection{Morphological Tasks}
Morphological tasks in NLP are  typically devided into \textit{generation} and \textit{analysis} tasks.
In both cases, the basic morphological structure assumed is an {\em inflection table}.
The dimensions of an inflection table are defined by a set of {\em attributes} (e.g., gender, number, case, etc.) and their possible {\em values} (e.g., gender:\{masculine,feminine,neuter\}).  
A specific {\em attribute:value} pair defines an {\em inflectional feature} (henceforth, a {\em feature}) and a specific combination of features is called an {\em inflectional feature bundle} (here, a {\em feature bundle}). An inflection table includes, for a given lexeme $l_i$, an exhaustive list of \(m\) inflected word-forms $\{w_{b_j}^{l_i}\}_{j=0}^m$,  corresponding to all available {\em feature bundles} $\{b_j\}_{j=0}^m$. See Table~\ref{tab:swa} for a fraction of an inflection table in Swahili. A {\em paradigm} in a language (verbal, nominal, adjectival, etc.) is a set of inflection tables. The set of inflection tables for a given language can be used to derive labeled data for (at least) 3  different tasks, {\em inflection, reinflection} and {\em analysis}.\footnote{The list of tasks mentioned above is of course not exhaustive; other tasks may be derived from labeled inflection tables, e.g., the Paradigm Cell Completion Problem \citep{ackerman-etal-2009-parts, cotterell-etal-2017-conll}.}

In  \textit{morphological inflection} (1a), the input  is a lemma \(l_i\) and a feature bundle \(b_j\) that specifies the target word-form. The output is the inflected word-form \(w_{b_j}^{l_i}\) realizing the feature bundle.  (1b) is an example  in the French verbal paradigm for the lemma \emph{finir}, 
inflected to an indicative \textsc{ind} future tense \textsc{fut}  with a 1st person singular subject \textsc{1;sg}.
\begin{enumerate}[nosep]
    \item[(1)] \begin{itemize}[nosep]
        \item[a.] 
\( \langle l_i, b_j \rangle \mapsto w_{b_j}^{l_i} \)
    \item[b.] \( \langle \emph{finir}, \textsc{ind;fut;1;sg} \rangle \mapsto \emph{finirai} \)
        \end{itemize}
\end{enumerate}

The morphological inflection task is in fact a specific version of a more general  task which is called  \textit{morphological reinflection}. In the general case, the source of inflection can be any form rather than only the lemma. Specifically, a source word-form \(w_{b_j}^{l_i}\) from some lexeme \(l_i\) is given as input accompanied by its own feature bundle \(b_j\), and the model reinflects it to a different %
feature bundle \(b_k\), resulting in the word  \(w_{b_k}^{l_i}\) (2a). In (2b) we illustrate for the same French lemma {\em finir},  a reinflection from the indicative present tense with a first person singular subject \emph{`finis'} to the subjunctive past and  second person singular \emph{`finisses'}.
\begin{enumerate}[nosep]
    \item[(2)]
    \begin{itemize}[nosep]
        \item[a.] \( \langle b_j, w_{b_j}^{l_i}\rangle , \langle b_k, \_\_\_ \rangle \mapsto w_{b_k}^{l_i} \)
    \item[b.] \begin{tabular}{r@{}}
                    \(\langle\) \textsc{ind;prs;1;sg}, \emph{finis} \(\rangle\), \\
                    \(\langle\) \textsc{sbjv;pst;2;sg}, \_\_\_ \(\rangle\)
                    \end{tabular}
                    \hphantom{~}$\mapsto$ \emph{finisses}
                \end{itemize}
\end{enumerate}

Morphological inflection and reinflection are generation tasks, in which word forms are generated from feature specifications.
In the opposite direction, \textit{morphological analysis} is a task where word-forms are the input, and models map them to their lemmas and feature bundles (3a). This task is in fact an inverted version of inflection, as can be seen in (3), which are the exact inverses of (1).
\begin{enumerate}[nosep]
\item[(3)]
\begin{itemize}[nosep]
    \item[a.] \( w_{b_j}^{l_i} \mapsto \langle l_i, b_j \rangle \)
    \item[b.] \( \emph{finirai} \mapsto \langle \emph{finir}, \textsc{ind;fut;1;sg} \rangle \)
    \end{itemize}
\end{enumerate}

\subsection{UniMorph}

\input{clause_tables}

The most significant source of inflection tables for training and evaluating   all of the aforementioned tasks is UniMorph\footnote{\url{https://unimorph.github.io}} \citep{sylak-glassman-etal-2015-language, batsuren-etal-2022-unimorph}, a large inflectional-morphology dataset covering over 170 languages. For each language the data contains a list of lexemes with all their associated feature bundles and the words realizing them.
Formally, every entry in UniMorph is a triplet \(\langle\)\textit{l,b,w}\(\rangle\) with \textit{lemma} \textit{l},  a \textit{feature bundle} \textit{b}, and  a \textit{word-form} \textit{w}.
The tables in UniMorph are exhaustive, that is, the data generally does not contain partial tables; their structure is fixed for all lexemes of the same paradigm, and each cell is filled in with a single form, unless that form doesn't exist in that language.\footnote{In cases of overabundance, i.e., availability of more than one form per cell, only one canonical form occupies the cell.} 
The data is usually crawled from Wiktionary\footnote{\url{https://www.wiktionary.org}} or from some preexisting finite-state automaton. The features for all languages are standardized to be from a shared inventory of features, but every language makes use of a different subset of that inventory. 
 
So far, the formal definition of UniMorph {\em seems}  cross-linguistically consistent. However, a closer inspection of UniMorph reveals an inconsistent definition of {\em words}, which then influences the dimensions included in the inflection tables in different languages. For example, the Finnish phrase \textit{`olen ajatellut'} is considered a single word, even though it contains a white-space. It is included in the relevant  inflection table and annotated as \textsc{act;prs;prf;pos;ind;1;sg}.  Likewise, the Albanian phrase \textit{`do të mendosh'} is also considered a single word, labeled as \textsc{ind;fut;1;pl}. In contrast, the English equivalents \textit{have thought} and \textit{will think}, corresponding to the exact same feature-bundles and meanings, are absent from UniMorph, and their construction is considered purely syntactic.

This overall inconsistency encompasses the inclusion or exclusion of various auxiliary verbs as well as the inclusion of  particles, clitics, light verb constructions and more.
The decision on what or how much phenomena to include is done in a per-language fashion that is inherited from the specific language's grammatical traditions and sources. In practice, it is quite arbitrary and taken without any consideration of universality. In fact, the definition of inflected words can be inconsistent even in closely related languages in the same language family, {e.g., the  Arabic definite article is included in the Arabic nominal paradigm,  while the equivalent definite article is excluded for Hebrew nouns.}

One possible attempted solution could be to define words by white-spaces and strictly exclude any forms with more than one space-delimited word. However, this kind of solution will severely impede the universality of any morphological task as it would give a tremendous weight to the orthographic tradition of a language and would be completely inapplicable for languages that do not use a word-delimiting sign like Mandarin Chinese and Thai.
On the other hand, a decades-long debate about a space-agnostic word definition have failed to result in any workable solution (see Section \ref{sec:related}).%

We therefore suggest to proceed in the opposite, far more inclusive, direction. We propose not to  try to delineate `words', but rather a consistent feature set to inflect lexemes for, regardless of the number of `words' and white spaces needed to realize it.%

\section{The Proposal: Word-Free Morphology}
\label{sec:clause_morph}

In this work we extend inflectional morphology, data and tasks, to the clause level. We define an inclusive cross-lingual set of inflectional features $\{b_j\}$ and inflect lemmas in all languages to the same set, no matter how many white-spaces have to be used in the realized form.
By doing so, we reintroduce universality into morphology, equating the treatment of languages in which clauses are frequently expressed with a single word with those that use several of them. Figure~\ref{fig:overview} exemplifies how this approach induces universal treatment for typologically different languages, as lexemes are inflected to the same feature bundles in all of them.

\paragraph{The Inflectional Features}
Our guiding principle in defining an inclusive set of features is the inclusion of all feature types expressed at word level in \textit{some} language. This set essentially defines a {\em saturated clause}. %

Concretely, our universal feature set contains the obvious {\em tense, aspect} and {\em mood} (TAM) features, as well as {\em negation, interrogativity} and all {\em argument-marking} features such as: {\em  person, number, gender, case, formality} and {\em reflexivity.} TAM features are obviously included as %
the hallmark of almost any inflectional system, particularly in most European languages, {\em negation} is expressed at the word level in many Bantu languages \citep{wilkes-nkosi-2012-complete, mpiranya-2014-swahili}, and {\em interrogativity} --- in, e.g., Inuit \citep{webster-1968-let} and to a lesser degree in Turkish.

Perhaps more important (and less familiar) is the fact that in many %
languages
multiple arguments can be marked on a single verb. For example, agglutinating languages  like Georgian and Basque show poly-personal agreement,
where the verb morphologically indicates features of {\em multiple} arguments, above and beyond the subject. For example:
 
\begin{enumerate}[nosep, leftmargin=1.4em]
    \item[(4)] \begin{itemize}[nosep, leftmargin=0.9em, labelsep=0.3em]
        \item[a.] Georgian: \kat{gagi+svebt}\\
        Trans: ``{\bf we} will let {\bf you} go"\\   \textsc{ind;fut;nom(1,pl);acc(2,sg)}
        \item[b.] Spanish: \textit{dímelo}\\
        Trans: ``tell {\bf it} to {\bf me}"\\\textsc{imp;nom(2,sg);acc(3,sg,neut);dat(1,sg)}
        \item[c.] Basque:  \textit{dakarkiogu}\\
        Trans: ``{\bf we} bring {\bf it} to {\bf him/her}"
        \\ \textsc{ind;prs;erg(1,pl);abs(3,sg);dat(3,sg)}
    \end{itemize}
\end{enumerate}
 
Following \citet{anderson-1992-morphous}'s  feature layering approach, 
we propose the annotation of arguments to be done as complex features, i.e., features that allow a feature set as their value.\footnote{This is reminiscent of feature structures in Unification Grammars \cite{shieber} such as GPSG, HPSG, LFG \cite{gpsg,pollard-sag-1994-head,bresnan-etal-2015-lexical}.}
So, the Spanish verb form \textit{dímelo},
(translated: `tell it to me'), for example, will be tagged as \textsc{imp;nom(2,sg);acc(3,sg,neut);dat(1,sg)}.

For languages that do not mark the verb's arguments by morphemes, we use personal pronouns to realize the relevant feature-bundles, e.g., the \textbf{bold} elements in the English translations in (4). %
Treating pronouns as feature realizations keeps the clauses single-lexemed for all languages, whether argument incorporating or not. To keep the inflected clauses single-lexemed in this work, we also limit the forms to main clauses,  avoiding subordination.

Although we collected the inflectional features %
empirically and bottom-up, the list we ended up with corresponds to \citet[p.~219]{anderson-1992-morphous}'s suggestion for clausal inflections: ``[for VP:] auxiliaries, tense markers, and pronominal elements representing the arguments of the clause; and determiners and possessive markers in NP''. %
Thus, our suggested feature set is not only  diverse and inclusive in practice, it is also theoretically sound.\footnote{Our resulted set may still be incomplete, but the principle holds: when adding a new language with new word-level features, these features will be %
realized for {all} languages.
}%

To illustrate, Table~\ref{tab:clause_table} shows a fragment of a clause-level inflection table in Swahili and its English equivalent. It shows that while the Swahili forms are expressed with one word, their English equivalents express the same feature bundles with several words. Including the exact same feature combinations, while allowing for multiple `word' expressions in the inflections, finally makes the comparison between the two languages straightforward, and showcases the comparable complexity  of {\em clause-level morphology} across languages.

\paragraph{The Tasks}
To formally complement our proposal, We  amend the task definitions in Section~\ref{sec:word_morph} to refer to forms in general $f_{b_j}^{l_i}$ rather than words  $w_{b_j}^{l_i}$ :

\begin{enumerate}[nosep]
    \item[(5)]  Clause-Level Morphological Tasks 
    \begin{itemize}[nosep]
    \item[a.]
    \makebox[2cm][l]{inflection} $ \langle l_i, b_j \rangle \mapsto f_{b_j}^{l_i} $
    \item[b.] \makebox[2cm][l]{reinflection} $ \langle b_j, f_{b_j}^{l_i}\rangle , \langle b_k, \_\_\_ \rangle \mapsto f_{b_k}^{l_i} $
    \item[c.] \makebox[2cm][l]{analysis} $ f_{b_j}^{l_i} \mapsto \langle l_i, b_j \rangle $
\end{itemize}
\end{enumerate}

See Table~\ref{tab:examples} for detailed examples of these tasks  for all the languages included in this work.

\input{examples}

\section{The \textsc{MightyMorph} Benchmark}
\label{sec:mightymorph}

We present \textsc{MightyMorph}, the first multilingual clause-level morphological dataset. Like UniMorph, \textsc{MightyMorph} contains inflection tables with entries of the form of \textit{lemma, features, form}. The data can be used to elicit training sets for any clause-level morphological task.

The data  covers four languages from three language families: English, German, Turkish and Hebrew.\footnote{For Hebrew, we annotated a vocalized version in addition to the commonly-used unvocalized forms.}
Our selection covers languages classified as isolating, agglutinative and fusional. The languages vary also in the extent they utilize  morpho-syntactic processes: from the ablaut extensive Hebrew to no ablauts in Turkish;  from fixed word order in Turkish to the meaning-conveying  word-order in German. Our data for each language contains at least 500 inflection tables.

Our data is currently limited to clauses constructed from verbal lemmas, as these are typical clause heads. Reserved for future work is the expansion of the process described below to nominal and adjectival clauses.

\subsection{Data Creation}
\label{sec:creation}

The data creation process, for any language, can be characterized by three conceptual components: (i) Lexeme Sampling, (ii) Periphrastic Construction, and (iii) Argument Marking. We describe each of the different phases in turn. As a running example, we will use %
the English verb \textit{receive}.

\paragraph{Lexeme Sampling.}
To create \textsc{MightyMorph}, we first sampled frequently used verbs from UniMorph. We assessed the verb usage by the position of the lemma in the frequency-ordered vocabulary of the FastText word vectors \cite{grave-etal-2018-learning}.\footnote{\url{https://fasttext.cc/docs/en/crawl-vectors.html}} We excluded auxiliaries and any lemmas frequent due to homonymy with non-verbal lexemes.%

\paragraph{Periphrastic Constructions}
We expanded each verb's word-level inflection table to include all periphrastic constructions using a language-specific rule-based grammar we wrote and the inflection tables of any relevant auxiliaries. I.e., we constructed forms for all possible TAM combinations expressible in the language, regardless of the number of words used to express this combination of features.
E.g., when constructing the future perfect form with a 3rd person singular subject for the lexeme \textit{receive}, equivalent to \textsc{ind;fut;prf;nom(3,sg)}, we used the past participle from the UniMorph inflection table \textit{received} and the auxiliaries \textit{will} and \textit{have} to construct \textit{will have received}.%

\paragraph{Argument Marking}
At first, we added the pronouns that the verb agrees with, unless a pro-drop applies. For all languages in our selection, the verb agrees only with its subject. A place-holder was then added to mark the linear position of the rest of the arguments. 
So the form of our example is now \textit{he will have received \fauxsc{args}}.

In order to obtain a fully-saturated clause, but also not to over-generate redundant arguments --- for example, a transitive clause for an intransitive verb --- an exhaustive list of \textit{frames} for each verb is needed. The frames are lists of cased arguments that the verb takes. For example the English verb \textit{receive} has 2 frames
\textsc{\{nom, acc\}} and \textsc{\{nom, acc, abl\}}, where an accusative argument indicates theme and the an ablative argument marks the source. When associating verbs with their arguments we did not restrict ourselves to the distinction between intransitive, transitive and ditransitive verbs,  we allow arguments of any case.
We treated all argument types equally and annotated them with a case feature, whether expressed with an affix, an adposition or a coverb. Thus, English \textit{from you}, Turkish \textit{senden} and Swahili \textit{kutoka kwako} are all tagged with an ablative case feature \textsc{abl(2,sg)}. 

For each frame we exhaustively generated all suitably cased pronouns without regarding the semantic plausibility of the resulted clause. So the clause \textit{he will have received you from it} is in the inflection table since it is grammatical -- even though it sounds odd. In contrast, \textit{he will have received} is not in the inflection table, as it is strictly ungrammatical, missing (at least) one obligatory argument.

Notably, we excluded adjuncts from the possible frames, defined here as argument-like elements that can be added to all verbs without regards to their semantics, like beneficiary and location.

We manually annotated 500 verbs in each language with a list of frames, each listing 0 or more arguments. This is the only part of the annotation process that required manual treatment of individual verbs.\footnote{ Excluding the manual work that may have been put in constructing the UniMorph inflection tables to begin with.} %
It was done by the authors, with the help of a native speaker or a monolingual dictionary.\footnote{%
For German we used \href{https://www.duden.de}{Duden dictionary}, and for Turkish we used the \href{https://sozluk.gov.tr}{Türk Dil Kurumu}'s dictionary.}

\input{stats_table}

We built an annotation framework that delineates the different stages of the process. It includes an infrastructure for grammar description and an interactive frame annotation component. Given a grammar description, the system handles the sampling procedure and constructs all relevant periphrastic constructions while leaving an additional-arguments place-holder. After receiving the frame-specific arguments from the user, the system completes the sentence by replacing the place holder with all prespecified pronouns for the frame.
The framework can be used to speed up the process of adding more languages to \textsc{MightyMorph}.\footnote{The data and annotation scripts are available at \url{https://github.com/omagolda/mighty_morph_tagging_tool}.} Using this framework, we have been able to annotate 
500 verb frames in about 10 hours per language on average.

\subsection{The Annotation Schema}
\label{sec:schema}

Just as our data creation builds on the word-level inflection tables of UniMorph and expands them, so our annotation schema is built upon UniMorph's.

In practice, due to the fact that some languages do use a single word for a fully-saturated clause, we could simply apply the UniMorph annotation guidelines \cite{sylak-2016-composition} both as an inventory of features and as general guidelines for the features' usage. Adhering to these guidelines ensures that our approach is able to cover essentially all languages covered by UniMorph. In addition, we extended the schema with the layering mechanism described in Section~\ref{sec:clause_morph} and by \citet{guriel-etal-2022-morphological}, and officially adopted as part of the UniMorph schema by \citet{batsuren-etal-2022-unimorph}.%

See Table~\ref{tab:features} for a detailed list of features used.

\begin{table*}[t]
    \centering
    \resizebox{\textwidth}{!}{
    \begin{tabular}{c|c|c}
    \hline
    \multicolumn{2}{c|}{\textbf{Attribute}}  &  \textbf{Value} \\
    \hline
    \hline
    \multicolumn{2}{c|}{Tense} &\textsc{pst}(past),\textsc{prs}(present),\textsc{fut}(future)\\
    \hline
    \multicolumn{2}{c|}{Mood} &
    \begin{tabular}[c]{@{}c@{}}\textsc{ind}(indicative)
    \textsc{imp}(imperative)
    \textsc{sbjv}(subjunctive)
    \textsc{infr}\textsuperscript{\textdagger}(inferential)\\ \textsc{nec}\textsuperscript{\textdagger}(necessitative)
    \textsc{cond}(conditional)
    \textsc{quot}(quotative)\end{tabular}\\
    \hline
    \multicolumn{2}{c|}{Aspect} & \textsc{hab}(habitual)
    \textsc{prog}(progressive)
    \textsc{prf}(perfect)
    \textsc{prsp}(prospective)\\
    \hline
    \multicolumn{2}{c|}{Non-locative Cases} &  \textsc{nom}(nominative)
    \textsc{acc}(accusative)
    \textsc{dat}(dative)
    \textsc{gen}(genitive)
    \textsc{com}(comitative)
    \textsc{ben}(benefactive)\\
    \hline
    \multicolumn{2}{c|}{Locative Cases} &
    \begin{tabular}[c]{@{}c@{}}
        \textsc{loc}\textsuperscript{\textdagger}(general locative)
        \textsc{abl}(ablative)
        \textsc{all}(allative)
        \textsc{ess}(essive)
        \textsc{apud}(apudessive)
        \textsc{perl}\textsuperscript{\textdagger}(perlative)\\
        \textsc{circ}(near)
        \textsc{ante}(in front)
        \textsc{contr}\textsuperscript{\textdagger}(against)
        \textsc{at}(at, general vicinity)
        \textsc{on}(on)
        \textsc{in}(in)
        \textsc{von}\textsuperscript{\textdagger}(about)
    \end{tabular}\\
    \hline
    \multicolumn{2}{c|}{Sentence Features} &  \textsc{neg}(negative)
    \textsc{q}(interrogative)\\
    \hline
    \multirow{4}{*}{\begin{tabular}[c]{@{}c@{}}Argument\\Features\end{tabular}} & Person & 1(1st person) 2(2nd person) 3(3rd person)\\
    \cline{2-3}
    & Number & \textsc{sg}(singular) \textsc{pl}(plural)\\
    \cline{2-3}
    & Gender & \textsc{masc}(masculine) \textsc{fem}(feminine) \textsc{neut}(neuter)\\
    \cline{2-3}
    & Misc. & \textsc{form}(formal) \textsc{rflx}\textsuperscript{\textdagger}(reflexive)

    \end{tabular}
    }
    \caption{A list of all features used in constructing the data for the 4 languages in \textsc{MightyMorph}. Upon addition of new languages the list would expand. 
    Features not taken from \citet{sylak-2016-composition}.
    are marked with \textdagger.}
    \label{tab:features}
\end{table*}

\subsection{Data Analysis}
\label{sec:data_stats}

The  {\sc MightyMorph} benchmark represents inflectional morphology in four typologically diverse languages, yet, the data is both more uniform across languages and more diverse in the features  realized for each language, compared to the de-facto standard word-level morphological annotations.

Table~\ref{tab:stats} compares aggregated values between UniMorph and \textsc{MightyMorph} across languages: the inflection table size,\footnote{Since the table size is dependent on the transitivity of the verb, the clause level is compared to an intransitive table.} %
the number of unique features used, the average number of features per form, and the average form-length in characters. 

We see that \textsc{MightyMorph} is more  cross-lingually consistent than UniMorph on all four comparisons: the size of the tables is less varied, so English no longer has extraordinarily small tables; 
the sets of features that were used per language are very similar, due to the fact that they all come from a fixed inventory; 
and finally, forms in all languages are of similar character length and are now described by feature bundles whose feature length are also highly similar. 
The residual variation in all of these values arises only from true linguistic variation.
For example, Hebrew does not use features for aspects as Hebrew does not express verbal aspect at all.
This is a strong empirical indication that applying morphological annotation to clauses  reintroduces universality into morphological data. 

In addition, the bigger inflection tables in \textsc{MightyMorph} include phenomena more diverse, %
like word-order changes in English, lexeme-dependent perfective auxiliary in German, and partial pro-drop in Hebrew. Thus, models trying to tackle clause-level morphology will need to address these newly added phenomena. We conclude that our proposed data and tasks are 
more universal than the previously-studied word-level morphology.

\section{Experiments}
\label{sec:experiments}

\input{results_tables}

\paragraph{Goal}
We set out to assess the challenges and opportunities presented to contemporary models by clause-level morphological tasks. To this end 
we experimented with the 3 tasks defined in Section~\ref{sec:clause_morph}: inflection, reinflection and analysis, all executed both at the word-level and the clause-level.

\paragraph{Splits}
For each task we sampled from 500 inflection tables 10,000 examples (pairs of examples in the case of reinflection). %
We used 80\% of the examples for training and the rest was divided between the validation and test sets. We sampled the same number of examples from each table and, following \citet{goldman-etal-2022-unsolving}, we split the data such that the lexemes in the different sets are disjoint. So, 400 lexemes are used in the train set, and 50 are for each of the validation and test sets.

\paragraph{Models}
As %
baselines, we applied contemporary models designed for word-level morphological tasks (henceforth: word-level models). 
The application of  word-level models will allow us to assess the difficulty of the clause-level tasks comparing to their word-level counterparts.
These models generally handle characters as input and output, and we applied them to  clause-level tasks straightforwardly by treating white-space as yet another character rather than a special delimiter. 
{For each language and task we trained a separate model for 50 epochs.} 
The word-level models  we trained are:
\begin{itemize}[nosep]
    \item \textsc{lstm}: 
    An LSTM encoder-decoder with attention, by \citet{silfverberg-hulden-2018-encoder}.
    \item \textsc{Transduce}: A neural transducer predicting actions between the input and output strings, by \citet{makarov-clematide-2018-neural}.
    \item \textsc{DeepSpin}: an RNN-based system using sparsemax instead of softmax, by \citet{peters-martins-2020-one}.
\end{itemize}
All models were developed for word-level {\em inflection}. \textsc{Transduce} is the SOTA for low-resourced morphological inflection \citep{cotterell-etal-2017-conll}, and \textsc{DeepSpin} is the SOTA in the general setting \citep{goldman-etal-2022-unsolving}.
We modified \textsc{Transduce} to apply to reinflection, %
while only the generally-designed \textsc{lstm} could be used for all tasks.

In contrast with word-level tasks, the extension of morphological tasks to the clause-level introduces {\em context} of a complete sentence, which provides an opportunity to explore the benefits of pre-trained {\em contextualized} LMs. success of such models on many NLP tasks calls for investigating their performance in our setting.
We thus used the following pretrained text-to-text model as an advanced modeling alternative for our clause-level tasks:
\begin{itemize}[nosep]
    \item \textsc{mT5}: 
    An encoder-decoder transformer-based model, pretrained by \citet{xue-etal-2021-mt5}
\end{itemize}
\textsc{mT5}'s input and output are tokens provided by the model's own tokenizer; the morphological features were used as a prompt and were added to the model's vocabulary as new tokens with randomly initialized embeddings.\footnote{As none of the models were designed to deal with hierarchical feature structures, the features' strings were flattened before training and evaluation.
For example, the bundle \textsc{ind;prs;nom(1,sg);acc(2,pl)} is replaced with \textsc{ind;prs;nom1;nomsg;acc2;accpl}.}

\subsection{Results and Analysis}

Table~\ref{tab:main_res} summarizes the results for all models and all tasks, for all languages. When averaged across languages, the results for the inflection task show a drop in performance for the word-inflection models (\textsc{lstm, DeepSpin} and \textsc{transduce}) on clause-level tasks, indicating that the clause-level task variants are indeed more challenging.
This pattern is even more pronounced in the results for the reinflection task which seems to be the most challenging clause-level task, presumably due to the need to identify the lemma in the sequence, in addition to inflecting it. In the analysis task, the only word-level model, \textsc{lstm}, actually performs better on the clause level than on the word level, but this seems to be the effect one outlier language, namely unvocalized Hebrew, where analysis models suffer from the lack of diacritization and extreme ambiguity.

Moving from words to clauses introduces context, and we hypothesized that this would enable contextualized pretrained LMs to shine. However, on all tasks \textsc{mT5} did not prove itself to be a silver bullet. That said, the strong pretrained model performed on par with the other models on the challenging reinflection task --- the only task involving complete sentences on both input and output --- in accordance with the model's pretraining.

In terms of languages, the performance of the word-level models seems correlated across languages, with notable under-performance over all tasks in German. In contrast, \textsc{mT5} seems to be somewhat biased towards the western languages, English and German, especially in the generation tasks, inflection and reinflection.

\begin{figure}[htbp]
    \centering
    \begin{subfigure}[t]{\columnwidth}
        \centering
        \resizebox{\columnwidth}{!}{
        \includegraphics{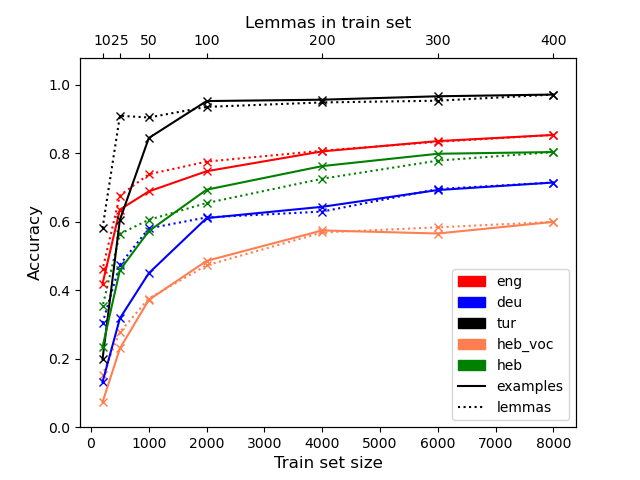}
        }
        \caption{Task: inflection; model: \textsc{transduce}}
        \label{fig:inf_curve}
    \end{subfigure}
    \begin{subfigure}[t]{\columnwidth}
        \centering
        \resizebox{\columnwidth}{!}{
        \includegraphics{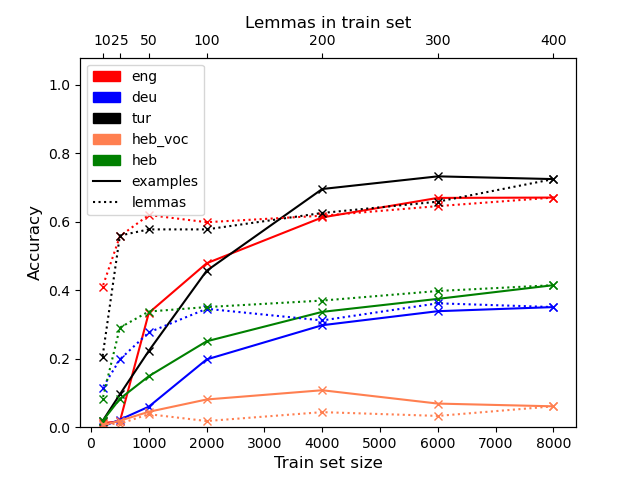}
        }
        \caption{Task: reinflection; model: \textsc{transduce}}
        \label{fig:reinf_curve}
    \end{subfigure}
    \begin{subfigure}[t]{\columnwidth}
        \centering
        \resizebox{\columnwidth}{!}{
        \includegraphics{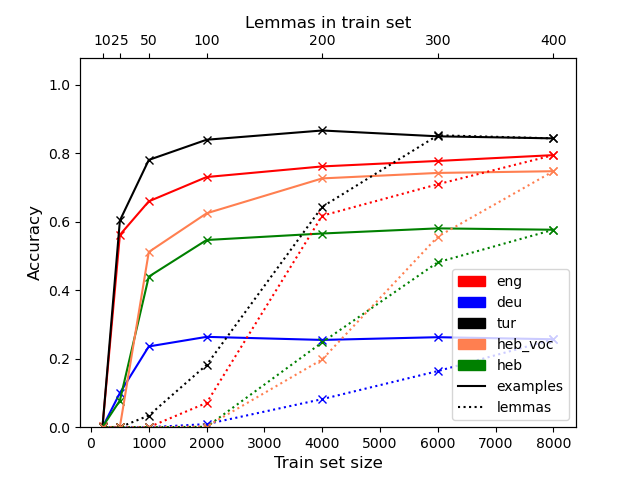}
        }
        \caption{Task: analysis; model: \textsc{lstm}}
        \label{fig:analysis_curve}
    \end{subfigure}
    \caption{Learning curves for the best performing model on each task. Solid lines are for increasing train set sizes while dashed lines --- for using more lexemes.}
    \label{fig:curves}
\end{figure}

\paragraph{Data Sufficiency}
To illustrate how much labeled data should {\em suffice} for training clause-morphology models, 
let us first note that the nature of morphology provides (at least) two ways to increase the amount of information available for the model. One is to increase the absolute number of sampled examples to larger training sets, while using the same number of inflection tables; alternatively, the number of inflection tables can be increased for a fixed size of the training set, increasing not the size but the variation in the set. 
The former is especially easy in languages with larger inflection tables, where each table can provide hundreds or thousands of inflected forms per lexeme, but the lack of variety in lexemes may lead to overfitting. To examine which dimension is more important for the overall success in the tasks, we tested both. 

The resulting curves are provided in Figure~\ref{fig:curves}.
In each sub-figure, the solid lines are for the results  as the absolute train set size is increased, and the dashed lines are for increasing the number of lexemes in the train set while keeping the absolute size of the train set fixed.

The resulting curves show that the balance between the options is different for each task. For inflection (Figure~\ref{fig:inf_curve}), increasing the size and the lexeme-variance of the training set produce similar trends, indicating that one dimension can compensate for the other. The curves for reinflection (Figure~\ref{fig:reinf_curve}) show that for this task the number of lexemes used is more important than the size of the training set, as the former produces steeper curves and reaches better performance with relatively little number of lexemes added. On the other hand, the trend for analysis (Figure~\ref{fig:analysis_curve}) is the other way around, with increased train set size being more critical than increased lexeme-variance.

\section{Related Work}
\label{sec:related}

\subsection{Wordhood in Linguistic Theory}
\label{sec:theory}

The quagmire surrounding words and their demarcation is long-standing in theoretical linguistics.  In fact, no coherent word definition has been provided by the linguistic literature despite many attempts. For example, \citet{zwicky-pullum-1983-cliticization} enumerate 6 different, sometimes contradictory, ways to discern between words, clitics and morphemes.
\citet{haspelmath-2011-indeterminacy} names 10 criteria for wordhood before concluding that no cross-linguistic definition of this notion can currently be found.

Moreover, words may be defined differently in different areas of theoretical linguistics. For example, the \textit{prosodic word} \citep{hall-1999-phonological} is defined in phonology and phonetics independently of the \textit{morphological word} \citep{bresnan-mchombo-1995-lexical}. And in general, many different notions of a word can be defined (e.g., \citealp{packard-2000-morphology} for Chinese).

However, the definition of morpho-syntactic words is inherently needed for the contemporary division of labour in theoretical linguistics, as it defines the boundary between morphology, the grammatical module in charge of word construction, and syntax, that deals with word combination \citep{dixon-aikhenvald-2002-word}. Alternative theories do exist, including ones that incorporate morphology into the syntactic constituency trees \citep{halle-marantz-1993-distributed}, and others that expand morphology to periphrastic constructions \citep{ackerman-stump-2004-paradigms} or to phrases in general \citep{anderson-1992-morphous}. In this work we follow that latter theoretical thread and expand morphological annotation up to the level of full clauses. This approach is theoretically leaner and requires less decisions that may be controversial, e.g., regarding morpheme boundaries, empty morphemes and the like.

The definition of words is also relevant to historical linguistics, where the common view considers items on a spectrum between words and affixes. Diachronically, items move mostly towards the affix end of the scale in a process known as \textit{grammaticalization} \citep{hopper-traugott-2003-grammaticalization} while occasional opposite movement is also possible \citep{norde-2009-degrammaticalization}. However, here as well it is difficult to find precise criteria for determining when exactly an item moved to another category on the scale, despite some extensive descriptions of the process (e.g., \citealp{joseph-2003-morphologization} for Greek future construction).

The vast work striving for cross-linguistically consistent definition of morpho-syntactic words seems to be extremely Western-biased, as it aspires to find a definition for words that will roughly coincide with those elements of text separated by white-spaces in writing of Western languages, rendering the endeavour particularly problematic for languages with orthographies that do not use white-spaces at all, like Chinese whose grammatical tradition contains very little reference to words up until the 20th century \citep{duanmu-1998-wordhood}. 

In this work we wish to bypass this theoretical discussion as it seems to lead to no workable word definition, and we therefore define morphology without the need of word demarcation.

\subsection{Wordhood in Language Technology}
\label{sec:practice}

The concept of words has been central to NLP from the very establishment of the field, as most models assume tokenized input (e.g., \citealp{richens-booth-1952-some,winograd-1971-procedures}). However, the lack of a word/token delimiting symbol in some languages prompted the development of more sophisticated tokenization methods, supervised \citep{xue-2003-chinese, nakagawa-2004-chinese} or statistical \citep{schuster-nakajima-2012-japanese}, mostly for east Asian languages.

Statistical tokenization methods also found their way to NLP of word-delimiting languages, albeit for different reasons like dealing with unattested words and unconventional spelling \citep{sennrich-etal-2016-neural, kudo-2018-subword}. Yet, tokens produced by these methods are sometimes assumed to correspond to linguistically defined units, mostly morphemes \citep{bostrom-durrett-2020-byte, hofmann-etal-2021-superbizarre}.

In addition, the usage of words as an organizing notion in theoretical linguistics, separating morphology from syntax, led to the alignment of NLP research according to the same subfields, with resources and models aimed either at syntactic or morphological tasks. For example, syntactic models usually take their training data from Universal Dependencies (UD; \citealp{demarneffe-etal-2021-universal}), where syntactic dependency arcs connect words as nodes while morphological features characterize the words themselves, although some works have experimented with dependency parsing of nodes other than words, be it chunks \citep{abney-1991-parsing, buchholz-etal-1999-cascaded} or nuclei \citep{barzdins-etal-2007-dependency, basirat-nivre-2021-syntactic}. However, in these works as well, the predicate-argument structure is still opaque in agglutinative languages where the entire structure is expressed in a single word.

Here we argue that questions regarding the correct granularity of input for NLP models will continue to haunt the research, at least until a thorough reference is made to the predicament surrounding these questions in theoretical linguistics. We proposed that given the theoretic state of affairs, a technologically viable word-free solution for computational morpho-syntax is desired, and this work can provide a stepping-stone for such a solution.

\section{Limitations and Extensions of Clause-Level Morphology}
\label{sec:discussion}

Our revised definition of morphology to disregard word boundaries does not (and is not intended to) solve all existing problems with  morphological annotations in NLP of course. Here we discuss some of the limitations and opportunities of this work for the future of morpho(syntactic) models in NLP.

\paragraph{The derivation-inflection divide.}
Our definition or clause-level morphology does not solve the long-debated demarcation of boundary between inflectional and derivational morphology (e.g., \citealp{scalise-1988-inflection}). Specifically, we only refered here to inflectional features, and, like UniMorph, did not provide a clear definition of what counts as inflectional vs.\ derivational. However, we suggest here that the lack of a clear boundary between inflectional and derivational morphology is highly similar to the lack of definition for words which operate as the boundary between morphology and syntax.
Indeed, in the theoretical linguistics literature, some advocate a view that posits {\em no} boundary between inflectional and derivational morphology \citep{bybee-1985-morphology}. Although this question is out of scope for this work, %
we conjecture that this similar problem may require a similar solution to ours, %
that will 
define a single framework for the entire inflectional--derivational morphology continuum without positing a boundary between them.

\paragraph{Overabundance.} Our shift to clause-level morphology does not solve the problem of overabundance, where several forms are occupying the same cell in the paradigm (for example, non-mandatory pro-drop in Hebrew). As the problem exists also in word-level morphology, we followed the same approach and constructed only one canonical form for each cell. However, for a greater empirical reach of our proposal, 
a further extension of the inflection table is conceivable, to accommodate sets of forms in every cell, rather than a single one.

\paragraph{Implications to syntax.} Our solution for annotating morphology at the clause level blurs the boundary between morphology and syntax as it is often presupposed in NLP, and thus has implications also for {\em syntactic} tasks. Some previous studies  indeed emphasized the cross-lingual inconsistency in word definition from the syntactic perspective \citep{basirat-nivre-2021-syntactic}. Our work  points to a holistic approach for morpho-syntactic annotation in which clauses are consistently tagged in a morphology-style annotation, leaving syntax for inter-clausal operations. %
Thus, we suggest that an extension of the approach taken here is desired in order to realize a single morpho-syntactic framework. Specifically, our approach should be extended to include: morphological annotation for clauses with multiple lexemes; realization of morphological features of  more clause-level characteristics, e.g., types of subordination and conjunction; and annotation of clauses in recursive structures. These are all fascinating research directions that extend the present contribution, and we reserve them for future work.

\paragraph{Polysynthetic languages.}
As a final note, we wish to make the observation that a unified morpho-syntactic system, whose desiderata are laid out in the previous paragraph, is essential for providing a straightforward treatment of some highly polysynthetic languages, specifically those that employ noun incorporation to regularly express some multi-lexemed clauses as a single word.

For example, consider the Yupik clause \textit{Mangteghangllaghyugtukut} translated \textit{We want to make a house}\footnote{Example adopted from Yupik UD, \citep{park-etal-2021-expanding}.}
containing 3 lexemes. Its treatment with the current syntactic tools is either non-helpful, as syntax only characterizes inter-word relations, or requires ad-hoc morpheme segmentation not used in other types of languages. Conversely, resorting to morphological tools will also provide no solution, due to the lexeme--inflection table paradigm that assumes single-lexemed words.
With a single morpho-syntactic framework, we could annotate the example above by incorporating the lemmas into their respective positions on the nested feature structure we used in this work, ending up with something similar to \textit{yug};\textsc{ind;erg(1;pl);comp}(\textit{ngllagh};\textsc{abs(-}\textit{mangtegha};\textsc{indef))}.
Thus, an annotation of this kind can expose the predicate-argument structure of the sentence while also being naturally applicable to other languages.

Equipped with these extensions, our approach could elegantly deal with polysynthetic languages and unlock a morpho-syntactic modeling ability that is most needed   for low-resourced languages.

\section{Conclusions}

In this work we expose the fundamental inconsistencies in contemporary computational morphology, namely, the inconsistency of {\em wordhood} across languages. To remedy this, we deliver \textsc{MightyMorph}, the first labeled dataset for clause-level morphology. We derive training  and evaluation data for the clause-level inflection, reinflection and analysis tasks. Our data analysis shows that the complexity of these tasks is more comparable across languages than their word-level counterparts. This reinforces our assumption that redefinition of morphology to the clause-level reintroduces universality into computational morphology.
Moreover, we showed that standard (re)inflection models struggle on the clause-level compared to their performance on word-level tasks, 
and that the challenge
is not trivially solved, even by contextualized pretrained  LMs such as \textsc{mT5}. %
In the future we intend to further expand our framework for more languages, and to explore more sophisticated models that take advantage of the hierarchical structure or better utilize pretrained LMs. Moreover, future work is planned to expand the proposal and benchmark to the inclusion of derivational morphology, and to a unified morpho-syntactic framework.

\section*{Acknowledgements}

We would like to thank the TACL anonymous reviewers and the action editor for their insightful suggestions and remarks. This work was supported funded by an ERC-StG grant from the  European Research Council, grant number 677352 (NLPRO), and by an innovation grant by the Ministry of Science and Technology (MOST) 0002214, for which we are grateful.

%% file: clause_tables.tex
\begin{table*}[t]
\begin{subtable}[h]{1\textwidth}
\centering
\scalebox{0.75}{
\begin{tabular}{l|cc|cc|cc}
\hline
\multirow{2}{*}{\begin{tabular}[c]{@{}l@{}}lexeme=PENDA\\ PRS;DECL;NOM(2,SG)\end{tabular}} & \multicolumn{2}{c|}{IND}                        & \multicolumn{2}{c|}{IND;PERF}                     & \multicolumn{2}{c}{COND}                             \\
                                                                                    & \multicolumn{1}{c|}{POS}          & NEG         & \multicolumn{1}{c|}{POS}          & NEG           & \multicolumn{1}{c|}{POS}           & NEG             \\ \hline
ACC(1,SG)                                                                           & \multicolumn{1}{c|}{unanipenda}   & hunipendi   & \multicolumn{1}{c|}{umenipenda}   & hujanipenda   & \multicolumn{1}{c|}{ungenipenda}   & usingenipenda   \\
ACC(1,PL)                                                                           & \multicolumn{1}{c|}{unatupenda}   & hutupendi   & \multicolumn{1}{c|}{umetupenda}   & hujatupenda   & \multicolumn{1}{c|}{ungetupenda}   & usingetupenda   \\
ACC(2,SG,RFLX)                                                                      & \multicolumn{1}{c|}{unajipenda}   & hujipendi   & \multicolumn{1}{c|}{umejipenda}   & hujajipenda   & \multicolumn{1}{c|}{ungejipenda}   & usingejipenda   \\
ACC(2,PL)                                                                           & \multicolumn{1}{c|}{unawapendeni} & huwapendini & \multicolumn{1}{c|}{umewapendeni} & hujawapendeni & \multicolumn{1}{c|}{ungewapendeni} & usingewapendeni \\
ACC(3,SG)                                                                           & \multicolumn{1}{c|}{unampenda}    & humpendi    & \multicolumn{1}{c|}{umempenda}    & hujampenda    & \multicolumn{1}{c|}{ungempenda}    & usingempenda    \\
ACC(3,PL)                                                                           & \multicolumn{1}{c|}{unawapenda}   & huwapendi   & \multicolumn{1}{c|}{umewapenda}   & hujawapenda   & \multicolumn{1}{c|}{ungewapenda}   & usingewapenda  \\
\hline
\end{tabular}
}
\caption{Swahili inflection table}
\label{tab:swa}
\end{subtable}

\vspace{0.3cm}

\begin{subtable}[h]{1\textwidth}
\centering
\scalebox{0.75}{
\begin{tabular}{l|cc|cc|cc}
\hline
\multirow{2}{*}{\begin{tabular}[c]{@{}l@{}}lexeme=LOVE\\ PRS;DECL;NOM(2,SG)\end{tabular}} & \multicolumn{2}{c|}{IND}                                                                                                                             & \multicolumn{2}{c|}{IND;PERF}                                                                                                                                 & \multicolumn{2}{c}{COND}                                                                                                                                        \\
                                                                                    & \multicolumn{1}{c|}{POS}                                                         & NEG                                                               & \multicolumn{1}{c|}{POS}                                                               & NEG                                                                  & \multicolumn{1}{c|}{POS}                                                                & NEG                                                                   \\ \hline
ACC(1,SG)                                                                           & \multicolumn{1}{c|}{you love me}                                                 & \begin{tabular}[c]{@{}c@{}}you don't \\ love me\end{tabular}      & \multicolumn{1}{c|}{\begin{tabular}[c]{@{}c@{}}you have\\ loved me\end{tabular}}       & \begin{tabular}[c]{@{}c@{}}you haven't\\ loved me\end{tabular}       & \multicolumn{1}{c|}{\begin{tabular}[c]{@{}c@{}}you would\\ love me\end{tabular}}       & \begin{tabular}[c]{@{}c@{}}you wouldn't\\ love me\end{tabular}       \\[0.3cm]
ACC(1,PL)                                                                           & \multicolumn{1}{c|}{you love us}                                                 & \begin{tabular}[c]{@{}c@{}}you don't\\ love us\end{tabular}       & \multicolumn{1}{c|}{\begin{tabular}[c]{@{}c@{}}you have\\ loved us\end{tabular}}       & \begin{tabular}[c]{@{}c@{}}you haven't\\ loved us\end{tabular}       & \multicolumn{1}{c|}{\begin{tabular}[c]{@{}c@{}}you would\\ love us\end{tabular}}       & \begin{tabular}[c]{@{}c@{}}you wouldn't\\ love us\end{tabular}       \\[0.3cm]
ACC(2,SG,RFLX)                                                                      & \multicolumn{1}{c|}{\begin{tabular}[c]{@{}c@{}}you love\\ yourself\end{tabular}} & \begin{tabular}[c]{@{}c@{}}you don't\\ love yourself\end{tabular} & \multicolumn{1}{c|}{\begin{tabular}[c]{@{}c@{}}you have\\ loved yourself\end{tabular}} & \begin{tabular}[c]{@{}c@{}}you haven't\\ loved yourself\end{tabular} & \multicolumn{1}{c|}{\begin{tabular}[c]{@{}c@{}}you would\\ love yourself\end{tabular}} & \begin{tabular}[c]{@{}c@{}}you wouldn't\\ love yourself\end{tabular} \\[0.3cm]
ACC(2,PL)                                                                           & \multicolumn{1}{c|}{you love y'all}                                              & \begin{tabular}[c]{@{}c@{}}you don't\\ love y'all\end{tabular}    & \multicolumn{1}{c|}{\begin{tabular}[c]{@{}c@{}}you have\\ loved y'all\end{tabular}}    & \begin{tabular}[c]{@{}c@{}}you haven't\\ loved y'all\end{tabular}    & \multicolumn{1}{c|}{\begin{tabular}[c]{@{}c@{}}you would\\ love y'all\end{tabular}}    & \begin{tabular}[c]{@{}c@{}}you wouldn't\\ love y'all\end{tabular}    \\[0.3cm]
ACC(3,SG)                                                                           & \multicolumn{1}{c|}{you love him}                                                & \begin{tabular}[c]{@{}c@{}}you don't\\ love him\end{tabular}      & \multicolumn{1}{c|}{\begin{tabular}[c]{@{}c@{}}you have\\ loved him\end{tabular}}      & \begin{tabular}[c]{@{}c@{}}you haven't\\ loved him\end{tabular}      & \multicolumn{1}{c|}{\begin{tabular}[c]{@{}c@{}}you would\\ love him\end{tabular}}      & \begin{tabular}[c]{@{}c@{}}you wouldn't\\ love him\end{tabular}      \\[0.3cm]
ACC(3,PL)                                                                           & \multicolumn{1}{c|}{you love them}                                               & \begin{tabular}[c]{@{}c@{}}you don't\\ love them\end{tabular}     & \multicolumn{1}{c|}{\begin{tabular}[c]{@{}c@{}}you have\\ loved them\end{tabular}}     & \begin{tabular}[c]{@{}c@{}}you haven't\\ loved them\end{tabular}     & \multicolumn{1}{c|}{\begin{tabular}[c]{@{}c@{}}you would\\ love them\end{tabular}}     & \begin{tabular}[c]{@{}c@{}}you wouldn't\\ love them\end{tabular}\\
\hline
\end{tabular}
}
\caption{English inflection table}
\label{tab:eng}
\end{subtable}

\caption{A fraction of a clause-level inflection table, in both English and Swahili. The tables are completely aligned in terms of meaning, but differ in the number of words needed to realize each cell. In practice, we did not inflect English clauses for number in 2nd person, so we did not use the \textit{y'all} pronoun and it is given here for the illustration.}
\label{tab:clause_table}
\end{table*}

%% file: examples.tex
\begin{table*}[t]
\begin{subtable}[h]{1\textwidth}
\centering
\scalebox{0.75}{
\begin{tabular}{l|c|c|c}
\hline
 & \multicolumn{2}{c|}{Input} & Output \\
 \hline
 \hline
 Eng          & give & \textsc{ind;fut;nom(1,sg);acc(3,sg,masc);dat(3,sg,fem)} & I will give him to her \\
\hline
Deu           & geben & \textsc{ind;fut;nom(1,sg);acc(3,sg,masc);dat(3,sg,fem)} & ich werde ihn ihr geben \\
\hline
Tur          & vermek & \textsc{ind;fut;nom(1,sg);acc(3,sg);dat(3,sg)} & onu ona vereceğim \\
\hline
Heb           & \cjRL{ntn} & \textsc{ind;fut;nom(1,sg);acc(3,sg,masc);dat(3,sg,fem)} & \cjRL{'tn 'wtw lh} \\
\hline
Heb$_{voc}$ & \cjRL{nAtan} & \textsc{ind;fut;nom(1,sg);acc(3,sg,masc);dat(3,sg,fem)} & \cjRL{'et*En 'OtO lAh*} \\
\hline
\end{tabular}
}
\caption{Inflection examples}
\label{tab:inf_ex}
\end{subtable}

\vspace{0.3cm}

\begin{subtable}[h]{1\textwidth}
\centering
\resizebox{\textwidth}{!}{
\begin{tabular}{l|c|c|c}
\hline
 & \multicolumn{2}{c|}{Input} & Output \\
 \hline
 \hline
\multirow{2}{*}{Eng} & \textsc{ind;fut;nom(1,sg);acc(3,sg,masc);dat(3,sg,fem)} & I will give him to her & \\
\cline{2-3}
& \textsc{ind;prs;nom(1,pl);acc(2);dat(3,pl);neg} & & we don't give you to them \\
\hline
\multirow{2}{*}{Deu} & \textsc{ind;fut;nom(1,sg);acc(3,sg,masc);dat(3,sg,fem)} & ich werde ihn ihr geben & \\
\cline{2-3}
& \textsc{ind;prs;nom(1,pl);acc(2,sg);dat(3,pl);neg} & & wir geben dich ihnen nicht \\
\hline
\multirow{2}{*}{Tur} & \textsc{ind;fut;nom(1,sg);acc(3,sg);dat(3,sg)} & onu ona vereceğim & \\
\cline{2-3}
& \textsc{ind;prs;prog;nom(1,pl);acc(2,sg);dat(3,pl);neg} & & seni onlara vermiyoruz \\
\hline
\multirow{2}{*}{Heb} & \textsc{ind;fut;nom(1,sg);acc(3,sg,masc);dat(3,sg,fem)} & \cjRL{'tn 'wtw lh} & \\
\cline{2-3}
& \textsc{ind;prs;nom(1,pl,masc);acc(2,sg,masc);dat(3,pl,fem);neg} & & \cjRL{'n.hnw l' nwtnym 'wtk lhn} \\
\hline
\multirow{2}{*}{Heb$_{voc}$} & \textsc{ind;fut;nom(1,sg);acc(3,sg,masc);dat(3,sg,fem)} & \cjRL{'et*En 'OtO lAh*} & \\
\cline{2-3}
& \textsc{ind;prs;nom(1,pl,masc);acc(2,sg,masc);dat(3,pl,fem);neg} & & \cjRL{'a:na.h:nw* lo' nwot:niym 'wot:kA lAhen} \\
\hline

\end{tabular}
}
\caption{Reinflection examples}
\label{tab:reinf_ex}
\end{subtable}

\vspace{0.3cm}

\begin{subtable}[h]{1\textwidth}
\centering
\scalebox{0.75}{
\begin{tabular}{l|c|c|c}
\hline
 & Input & \multicolumn{2}{c}{Output} \\
 \hline
 \hline
 Eng           & I will give him to her & give & \textsc{ind;fut;nom(1,sg);acc(3,sg,masc);dat(3,sg,fem)} \\
\hline
Deu            & ich werde ihn ihr geben & geben & \textsc{ind;fut;nom(1,sg);acc(3,sg,masc);dat(3,sg,fem)} \\
\hline
Tur           & onu ona vereceğim & vermek & \textsc{ind;fut;nom(1,sg);acc(3,sg);dat(3,sg)} \\
\hline
Heb            & \cjRL{'tn 'wtw lh} & \cjRL{ntn} & \textsc{ind;fut;nom(1,sg);acc(3,sg,masc);dat(3,sg,fem)} \\
\hline
Heb$_{voc}$ & \cjRL{'et*En 'OtO lAh*} & \cjRL{nAtan} & \textsc{ind;fut;nom(1,sg);acc(3,sg,masc);dat(3,sg,fem)} \\
\hline
\end{tabular}
}
\caption{Analysis examples}
\label{tab:analysis_ex}
\end{subtable}

\caption{Examples for the data format used for the inflection, reinflection ans analysis tasks.}
\label{tab:examples}
\end{table*}

%% file: stats_table.tex
\begin{table}[t]
\small{
\begin{center}

\resizebox{\columnwidth}{!}{
\begin{tabular}{l|cc|cc|cc|cc}
\hline
& \multicolumn{2}{c|}{Table size} & \multicolumn{2}{c|}{Feat set size} & \multicolumn{2}{c|}{Feats per form} & \multicolumn{2}{c}{Form length} \\
& UM & MM & UM & MM & UM & MM & UM & MM \\
 \hline
 \hline
\textbf{Eng} & 5 & 450 & 6 & 32 & 2.8 & 12.75 & 6.84 & 29.63 \\
\textbf{Deu} & 29 & 512 & 12 & 43 & 4.62 & 12.67 & 9.18 & 31.28 \\
\textbf{Heb} & 29 & 132 & 13 & 25 & 4.46 & 13.55 & 5.20 & 20.47 \\
\textbf{Heb}$_{voc}$ & 29 & 132 & 13 & 25 & 4.46 & 13.55 & 9.80 & 32.02 \\
\textbf{Tur} & 703 & 702 & 25 & 30 & 7.87 & 11.95 & 17.81 & 28.71 \\
\hline
\end{tabular}}
\end{center}
}
\caption{Comparison of statistics over the 4 languages common to UniMorph (UM) and \textsc{MightyMorph} (MM).
In all cases, the values for \textsc{MightyMorph} are more uniform across languages.}
\label{tab:stats}
\end{table}

%% file: results_tables.tex
\begin{table*}[t]
 \begin{center}
\resizebox{\textwidth}{!}{
 \begin{tabular}{ll|c|c||c|c|c|c|c|c|c|c|c|c}
 \hline
  &  & \multicolumn{2}{c||}{Average} & \multicolumn{2}{c|}{Eng} & \multicolumn{2}{c|}{Deu} & \multicolumn{2}{c|}{Heb} & \multicolumn{2}{c|}{Heb\(_\textit{vocalized}\)} & \multicolumn{2}{c}{Tur} \\
 & & word & clause & word & clause & word & clause & word & clause & word & clause & word & clause \\
 \hline
 \hline
 
 \multirow{4}{*}[-0.5cm]{inflec.} & \textsc{lstm} & \begin{tabular}[c]{@{}c@{}}84.7\\ $\pm$1.1\end{tabular} & \begin{tabular}[c]{@{}c@{}}70.0\\ $\pm$1.2\end{tabular} &  \begin{tabular}[c]{@{}c@{}}86.0\\ $\pm$1.8\end{tabular} & \begin{tabular}[c]{@{}c@{}}68.5\\ $\pm$3.8\end{tabular} & \begin{tabular}[c]{@{}c@{}}64.5\\ $\pm$4.7\end{tabular} & \begin{tabular}[c]{@{}c@{}}47.5\\ $\pm$4.0\end{tabular} & \begin{tabular}[c]{@{}c@{}}90.7\\ $\pm$1.6\end{tabular} & \begin{tabular}[c]{@{}c@{}}82.5\\ $\pm$0.6\end{tabular}& \begin{tabular}[c]{@{}c@{}}91.7\\ $\pm$1.1\end{tabular} & \begin{tabular}[c]{@{}c@{}}\textbf{70.0}\\ \textbf{$\pm$1.2}\end{tabular} & \begin{tabular}[c]{@{}c@{}}90.8\\ $\pm$0.9\end{tabular} & \begin{tabular}[c]{@{}c@{}}81.6\\ $\pm$2.1\end{tabular} \\ 
  \cline{2-14}
& \textsc{DeepSpin} & \begin{tabular}[c]{@{}c@{}}89.4\\ $\pm$0.8\end{tabular} & \begin{tabular}[c]{@{}c@{}}71.8\\ $\pm$0.5\end{tabular} & \begin{tabular}[c]{@{}c@{}}87.3\\ $\pm$2.8\end{tabular} & \begin{tabular}[c]{@{}c@{}}78.4\\ $\pm$1.5\end{tabular} & \begin{tabular}[c]{@{}c@{}}78.2\\ $\pm$0.5\end{tabular} & \begin{tabular}[c]{@{}c@{}}40.0\\ $\pm$0.5\end{tabular} & \begin{tabular}[c]{@{}c@{}}90.9\\ $\pm$0.2\end{tabular} & \begin{tabular}[c]{@{}c@{}}\textbf{86.1}\\ \textbf{$\pm$0.7}\end{tabular}& \begin{tabular}[c]{@{}c@{}}93.1\\ $\pm$2.0\end{tabular} & \begin{tabular}[c]{@{}c@{}}\textbf{71.7}\\ \textbf{$\pm$0.7}\end{tabular} & \begin{tabular}[c]{@{}c@{}}97.5\\ $\pm$2.1\end{tabular} & \begin{tabular}[c]{@{}c@{}}82.7\\ $\pm$1.6\end{tabular} \\ 
\cline{2-14}
& \textsc{transduce} & \begin{tabular}[c]{@{}c@{}}86.7\\ $\pm$0.5\end{tabular} & \begin{tabular}[c]{@{}c@{}}\textbf{78.9}\\ \textbf{$\pm$0.4}\end{tabular} & \begin{tabular}[c]{@{}c@{}}86.8\\ $\pm$0.4\end{tabular} & \begin{tabular}[c]{@{}c@{}}\textbf{85.4}\\ \textbf{$\pm$1.1}\end{tabular} & \begin{tabular}[c]{@{}c@{}}76.6\\ $\pm$2.5\end{tabular} & \begin{tabular}[c]{@{}c@{}}\textbf{71.5}\\ \textbf{$\pm$1.3}\end{tabular} & \begin{tabular}[c]{@{}c@{}}89.4\\ $\pm$0.6\end{tabular} & \begin{tabular}[c]{@{}c@{}}80.4\\ $\pm$0.8\end{tabular}& \begin{tabular}[c]{@{}c@{}}81.1\\ $\pm$0.5\end{tabular} & \begin{tabular}[c]{@{}c@{}}60.0\\ $\pm$1.1\end{tabular} & \begin{tabular}[c]{@{}c@{}}99.4\\ $\pm$0.1\end{tabular} & \begin{tabular}[c]{@{}c@{}}\textbf{97.2}\\ \textbf{$\pm$0.5}\end{tabular} \\ 
\cline{2-14}
& \textsc{mT5} & NA & \begin{tabular}[c]{@{}c@{}}51.9\\ $\pm$1.1\end{tabular} & NA & \begin{tabular}[c]{@{}c@{}}70.7\\ $\pm$1.7\end{tabular} & NA & \begin{tabular}[c]{@{}c@{}}57.7\\ $\pm$3.3\end{tabular} & NA & \begin{tabular}[c]{@{}c@{}}48.0\\ $\pm$3.3\end{tabular}& NA & \begin{tabular}[c]{@{}c@{}}34.2\\ $\pm$1.4\end{tabular} & NA & \begin{tabular}[c]{@{}c@{}}48.7\\ $\pm$1.7\end{tabular} \\ 

\hline
\\[-0.2cm]
\hline
 
\multirow{3}{*}[-0.4cm]{reinflec.} & \textsc{lstm} & \begin{tabular}[c]{@{}c@{}}73.2\\ $\pm$1.6\end{tabular} & \begin{tabular}[c]{@{}c@{}}\textbf{45.4}\\ \textbf{$\pm$9.4}\end{tabular} & \begin{tabular}[c]{@{}c@{}}78.2\\ $\pm$6.3\end{tabular} & \begin{tabular}[c]{@{}c@{}}62.7\\ $\pm$2.5\end{tabular} & \begin{tabular}[c]{@{}c@{}}53.5\\ $\pm$3.5\end{tabular} & \begin{tabular}[c]{@{}c@{}}31.0\\ $\pm$1.7\end{tabular} & \begin{tabular}[c]{@{}c@{}}68.4\\ $\pm$1.6\end{tabular} & \begin{tabular}[c]{@{}c@{}}30.6\\ $\pm$29.8\end{tabular}& \begin{tabular}[c]{@{}c@{}}80.7\\ $\pm$1.9\end{tabular} & \begin{tabular}[c]{@{}c@{}}\textbf{31.4}\\ \textbf{$\pm$36.4}\end{tabular} & \begin{tabular}[c]{@{}c@{}}85.2\\ $\pm$2.2\end{tabular} & \begin{tabular}[c]{@{}c@{}}\textbf{71.1}\\ \textbf{$\pm$1.2}\end{tabular} \\ 
 \cline{2-14}
& \textsc{transduce} & \begin{tabular}[c]{@{}c@{}}75.1\\ $\pm$0.5\end{tabular} & \begin{tabular}[c]{@{}c@{}}\textbf{44.5}\\ \textbf{$\pm$0.8}\end{tabular} & \begin{tabular}[c]{@{}c@{}}82.7\\ $\pm$1.1\end{tabular} & \begin{tabular}[c]{@{}c@{}}67.1\\ $\pm$0.4\end{tabular} & \begin{tabular}[c]{@{}c@{}}81.5\\ $\pm$0.5\end{tabular} & \begin{tabular}[c]{@{}c@{}}35.5\\ $\pm$0.3\end{tabular} & \begin{tabular}[c]{@{}c@{}}77.2\\ $\pm$1.2\end{tabular} & \begin{tabular}[c]{@{}c@{}}\textbf{41.5}\\ \textbf{$\pm$2.2}\end{tabular}& \begin{tabular}[c]{@{}c@{}}49.2\\ $\pm$1.5\end{tabular} & \begin{tabular}[c]{@{}c@{}}6.1\\ $\pm$1.8\end{tabular} & \begin{tabular}[c]{@{}c@{}}84.7\\ $\pm$1.1\end{tabular} & \begin{tabular}[c]{@{}c@{}}\textbf{72.5}\\ \textbf{$\pm$2.7}\end{tabular} \\ 
 \cline{2-14}
& \textsc{mT5} & NA & \begin{tabular}[c]{@{}c@{}}\textbf{45.2}\\ \textbf{$\pm$1.8}\end{tabular} & NA & \begin{tabular}[c]{@{}c@{}}\textbf{73.6}\\ \textbf{$\pm$3.1}\end{tabular} & NA & \begin{tabular}[c]{@{}c@{}}\textbf{54.2}\\ \textbf{$\pm$2.0}\end{tabular} & NA & \begin{tabular}[c]{@{}c@{}}30.8\\ $\pm$4.2\end{tabular}& NA & \begin{tabular}[c]{@{}c@{}}\textbf{29.7}\\ \textbf{$\pm$1.9}\end{tabular} & NA & \begin{tabular}[c]{@{}c@{}}37.5\\ $\pm$7.0\end{tabular} \\ 

 \hline
 \\[-0.2cm]
 \hline

\multirow{2}{*}[-0.15cm]{analysis} & \textsc{lstm} & \begin{tabular}[c]{@{}c@{}}62.0\\ $\pm$0.9\end{tabular} & \begin{tabular}[c]{@{}c@{}}\textbf{64.4}\\ \textbf{$\pm$1.1}\end{tabular} & \begin{tabular}[c]{@{}c@{}}81.6\\ $\pm$0.4\end{tabular} & \begin{tabular}[c]{@{}c@{}}\textbf{79.5}\\ \textbf{$\pm$2.2}\end{tabular} & \begin{tabular}[c]{@{}c@{}}34.8\\ $\pm$2.2\end{tabular} & \begin{tabular}[c]{@{}c@{}}25.7\\ $\pm$0.6\end{tabular} & \begin{tabular}[c]{@{}c@{}}34.6\\ $\pm$1.3\end{tabular} & \begin{tabular}[c]{@{}c@{}}\textbf{57.7}\\ \textbf{$\pm$1.4}\end{tabular}& \begin{tabular}[c]{@{}c@{}}73.3\\ $\pm$3.6\end{tabular} & \begin{tabular}[c]{@{}c@{}}\textbf{74.8}\\ \textbf{$\pm$2.2}\end{tabular} & \begin{tabular}[c]{@{}c@{}}85.6\\ $\pm$0.7\end{tabular} & \begin{tabular}[c]{@{}c@{}}\textbf{84.4}\\ \textbf{$\pm$4.5}\end{tabular} \\ 
\cline{2-14}
& \textsc{mT5} & NA & \begin{tabular}[c]{@{}c@{}}42.8\\ $\pm$1.2\end{tabular} & NA & \begin{tabular}[c]{@{}c@{}}69.0\\ $\pm$1.2\end{tabular} & NA & \begin{tabular}[c]{@{}c@{}}\textbf{45.1}\\ \textbf{$\pm$2.7}\end{tabular} & NA & \begin{tabular}[c]{@{}c@{}}48.0\\ $\pm$3.3\end{tabular}& NA & \begin{tabular}[c]{@{}c@{}}34.2\\ $\pm$1.4\end{tabular} & NA & \begin{tabular}[c]{@{}c@{}}46.0\\ $\pm$4.5\end{tabular} \\ 
 \hline
 \end{tabular}
 }
 \end{center}
 
\caption{Word and clause results for all tasks, models and languages, stated in terms of exact match accuracy in percentage. Over clause tasks, for every language and task the best performing system is in \textbf{bold}, in cases 
that are too close to call, in terms of standard deviations, 
all best systems are marked. Results are averaged over 3 runs with different initializations and training data order.}
\label{tab:main_res}
\end{table*}